# Neural encoding and interpretation for high-level visual cortices based on fMRI using image caption features

Kai Qiao,[1] Chi Zhang,[1] Jian Chen,[1] Linyuan Wang,[1] Li Tong,[1] Bin Yan[1*]

[1] Academy of information systems engineering, PLA strategy support force information engineering university, Zhengzhou, 4512001, China.

*Correspondence: ybspace@hotmail.com

## Abstract

On basis of functional magnetic resonance imaging (fMRI), researchers are devoted to designing visual encoding models to predict the neuron activity of human in response to presented image stimuli and analyze inner mechanism of human visual cortices. Deep network structure composed of hierarchical processing layers forms deep network models by learning features of data on specific task through big dataset. Deep network models have powerful and hierarchical representation of data, and have brought about breakthroughs for visual encoding, while revealing hierarchical structural similarity with the manner of information processing in human visual cortices. However, previous studies almost used image features of those deep network models pre-trained on classification task to construct visual encoding models. Except for deep network structure, the task or corresponding big dataset is also important for deep network models, but neglected by previous studies. Because image classification is a relatively fundamental task, it is difficult to guide deep network models to master high-level semantic representations of data, which causes into that encoding performance for high-level visual cortices is limited. In this study, we introduced one higher-level vision task: image caption (IC) task and proposed the visual encoding model based on IC features (ICFVEM) to encode voxels of high-level visual cortices. Experiment demonstrated that ICFVEM obtained better encoding performance than previous deep network models pre-trained on classification task. In addition, the interpretation of voxels was realized to explore the detailed characteristics of voxels based on the visualization of semantic words, and comparative analysis implied that high-level visual cortices behaved the correlative representation of image content, because IC task pays more attention to discovering relationships of objects for higher-level semantic understanding of input images.

## Introduction

In neuroscience, researchers have been interested in how human visual cortices process information of visual scene to understand external world. Since functional magnetic resonance imaging (fMRI) [1] provides an effective tool to reflect neuron activity in visual cortices for presented image stimuli, researchers have been trying constructing visual encoding models [2, 3] to predict fMRI patterns in visual cortices. The predicting accuracy of fMRI patterns is used to evaluate the quality of the constructed models. Meanwhile, experiment results and analyzation based on the constructed models [4, 5] can help obtain new or validate previous understanding of visual cortices. In addition, visual encoding models can also contribute to visual decoding models [6-10] predicting stimuli information from fMRI patterns, because the two models can be associated through Bayesian theory.

The quantity of fMRI dataset is small compared to big dataset [11, 12] in computer vision domain, because collecting enormous fMRI dataset is time consuming and the comfort of participant subjects need to be guaranteed in the fMRI experiment. Currently, the two-step linearizing encoding [3] is the main manner to design visual encoding models. The linearizing encoding models commonly perform nonlinear feature representation of input image stimulus, and linearly map image features into each fMRI voxel based on a linear regression model. The linear mapping makes it easy to analyse the relationship between image features and visual cortices. Furthermore, some characteristics of visual cortices can be revealed through interpreting or visualizing features. Hence, linearizing models have been widely used and recognized in visual encoding domain. Thereinto, the nonlinear feature representation [13] of images plays an important role in visual encoding models, and one well-matched feature representation with visual cortices can improve the final encoding performance.

In these years, inspired at some human visual mechanism, some human-like structures [14] were proposed to finish various computer vision task. Meanwhile, researchers in neuroscience mainly depended on nonlinear representations [15, 16] of images in computer vision to construct visual encoding models. Studies in computer vision domain can be divided into two periods. In the era of hand-crafted designed features [17] for computer vision task, Kay et al. [18] employed gabor wavelet pyramid (GWP) [19] features to encode fMRI voxels of primary visual cortices, and Huth et al. [4] employed hand-marked semantic labels to encode fMRI voxels of high-level visual cortices. After computer vision domain stepped in the deep learning era [20], deep network models with powerful representation of images started to attract much attention in visual encoding domain.

Hierarchical features of deep network models were introduced to design many visual encoding models [21]. Improved encoding performance and new understanding of human cortices [22-25] were obtained through deep network models. Meanwhile, it also has proved that the hierarchical deep network structure better accords with hierarchical information processing in human visual cortices. However, there is still much difference in accurately predicting fMRI voxels [8, 25] in response to image stimuli, especially for those voxels in high-level visual cortices. The phenomenon leads to a query for the deep network structure.

Next, we tried to analyze the disadvantage of currently employed deep network models and introduced our viewpoint for the query. We found that those deep network models used for visual encoding were almost pre-trained on image classification task, except few unsupervised deep network models [26]. However, many elements including deep network structure, task, training techniques, and super parameters have an influence on the final feature representation of produced deep network models [20]. As we all know, deep network and big data are regarded as the two key elements to promote the deep learning age. Although various deep network structures [27, 28] have been proved effective for visual encoding, the role of dataset or task was usually neglected. In this case, what does only those deep network models pre-trained on image classification task mean for visual encoding? For low-level visual cortices, deep network models driven by the image classification task can also extract low-level features of images such as edges, texture through the low and middle layers, which has been proved in many previous studies through visualization of features [29, 30]. Thus, current deep network models are enough for encoding low-level visual cortices. However, for high-level visual cortices, those deep network models driven by the image classification task seem to be relatively limited. Because image classification is a basic task in the field of computer vision, and needs to identify categories of the main targets for input images. Therefore, the last several layers of features of deep network model driven pre-trained on the task, especially last fully connected layer of features, tend to focus on the main goals of input images, while ignoring some other features about background, other tiny objects. The kind of attention makes it easy to obtain accurate classification results but makes it difficult to understand global semantics of input images. The reality can be also proved by some work about saliency maps [31, 32] for deep network. On the other hand, human can grasp all key elements instead of main objects to obtain the relationship of different elements and higher-level understanding of images when viewing images. Thus, we assumed that current deep network models pre-trained on image classification are not enough for encoding voxels of high-level visual cortices.

Therefore, based on the hierarchical deep network structure, this study thinks of introducing more semantic recognition task to design the computation model for high-level visual cortices. Image caption (IC) task [33] aimed at predicting the corresponding description given one image started to attract attention, along with the performance of image classification task is at bottleneck. Compared to image classification, IC task is a kind of higher-level task and can drive the deep network structure to learn more semantic features of images, during which, not only main objects but also other global information including background, relationship between objects, and so on are utilized. Therefore, these features can be more potential for high-level visual cortices. In this study, we introduced one deep network model pre-trained on IC task to design the visual encoding model for high-level visual cortices. Experiments on three subjects demonstrated that the task should be considered while designing visual encoding models, in addition to deep network architecture.

In this study, our main contributions are as follows: 1) from the perspective of task (one driving force of deep network), we analyzed that current deep network models pre-trained on image classification are not enough for encoding high-level visual cortices; 2) we designed the image caption (IC) features based visual encoding model (ICFVEM) to use more semantic features to better encode voxels of high-level visual cortices; 3) we revealed the correlative representation of different image objects in high-level visual cortices; 4) we interpreted voxels of high-level visual cortices based on visualization of words and related voxels with semantic words; and 5) we gave rise to the future direction by considering the other deep learning element in addition to deep network structure.

## Materials and Methods

### Experimental data

The dataset employed in this study is from the previous study [34]. The dataset includes visual stimuli and corresponding fMRI data for three subjects, and is called BOLD51200. Image stimuli are selected from three popular dataset (SUN, COCO, and ImageNet) in computer vision domain. In total, 4916 unique images including 1916 ImageNet images, 2000 COCO images and 1000 SUN images, were included for the fMRI experiment. Compared to the other public fMRI dataset, the image stimuli are extremely diverse and overlapping with existing computer vision datasets. The fMRI dataset includes four high-level visual (HV) regions: Parahippocampal Place Area (PPA), Retrosplenial Complex (RSC), Occipital Place Area (OPA), and lateral occipital cortices (LOC), in addition to the early visual regions (EV). Each visual region was divided into left and right sub-regions, respectively. In the experiment, 4803 images were presented once, 112 images were repeatedly presented four times, and one image were repeatedly presented three times. In our experiment, we took those images that were presented only once for the subject as the training set, and those images that were presented repetitively three or four times as the testing set for each subject. For those repetitive images, the corresponding values of voxels are averaged. Finally, the training set and testing set includes 4803 and 113 samples, respectively. The detailed information about the visual stimuli and fMRI data in the BOLD51200 can be found in the previous studies [34]. The employed dataset is open source and can be freely downloaded from https://figshare.com/articles/BOLD51200/6459449.

**The overview of proposed method**

According to the above analysis in the introduction section, we assume that the deep network models driven by higher-level task can be considered to design the computation model for visual encoding. Hence, we introduced the image caption (IC) task that needs to predict one words of description for given image. In this way, the deep network model pre-trained on image caption task can extract more semantic features compared to the previous deep network models pre-trained on image classification task. Therefore, it can improve the encoding performance of high-level visual cortices. As shown in the Figure 1, previous deep network model pre-trained on classification task only predict the category of main object and obtains "bike" through the decoder (eg. several fully connected layers or global pooling operation) for the input image. In this way, the information about the "man", the "bike", and their relative location and relationship is neglected, which is difficult to form global semantics. In contrast, the deep network model pre-trained on IC task called ICM employed recurrent neural network (RNN) operation based on traditional CNN features to extract higher-level features, which included more semantic information about the global scene and was used to describe the input image through the decoder. Hence, the processed features called ICF were more semantic than CNN features and were suitable for high-level visual cortices. In this study, we proposed the ICF-based visual encoding model (ICFVEM) according to the linearizing encoding manner. Firstly, we employed ICM to extract ICF for each image stimulus in training set. Secondly, for each voxel in visual cortices, we trained one linear regression model to map ICF to the voxel. Finally, given one new image stimulus in testing set, the corresponding voxel responses can be predicted based on ICFVEM including one ICM and trained linear regression models. In next subsection, we will describe ICFVEM in detail.

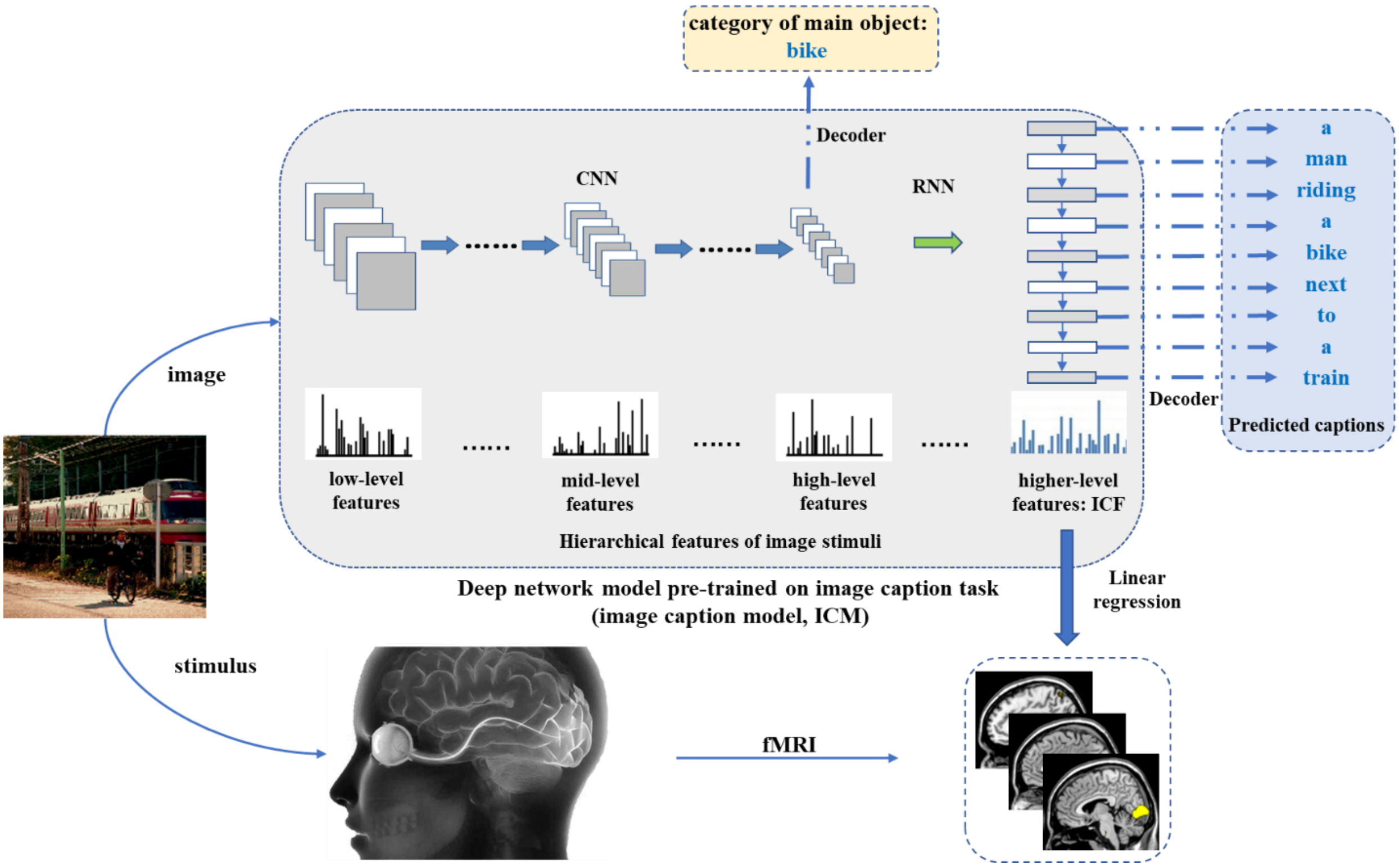

Figure 1. The image caption features based visual encoding model (ICFVEM). Visual encoding model is aimed at predicting fMRI patterns of viewed images based on image features. Higher-level visual cortices accords well with higher-level image features of images. The gray part represents the ICM including hierarchical CNN module and RNN module. The pale yellow part and the light blue part represents the image classification and caption task, respectively. Compared to image classification task that only needs to predict the category of main object (the "train") for input image, the image caption task requires to describe the input image using nearly all objects (the "man", the "train", the "bike") and their relationship (the man "rides" the bike "next to" the train), which produces the global semantic understanding for the image.

## Image caption features (ICF) based on image caption model (ICM)

In this study, we employed the ICM designed by [35] for image caption task to extract ICF of input images for visual encoding. ICM was composed of an encoder-decoder architecture and used to predict the caption of images. ICM firstly used an encoder to encode input image into low-dimensionality of features that included necessary information about all image content. Secondly, ICM used a decoder to decode a sequence of words as the predicted caption. Because convolution neural networks are suitable for representation learning of spatial images, the classic residual network (ResNet) [16] including 101 layers pre-trained on the ImageNet classification task was used as the encoder. Note that the extracted features based on the encoder are called CNN features in this study. Because recurrent neural networks (RNNs) are suitable for sequence modeling, one RNN module is used as the decoder to generate the sequence of words. Overall model were trained for image caption task in an end-to-end manner. Because ICF is constructed through the RNN decoder, hence, we next mainly introduce the RNN decoder.

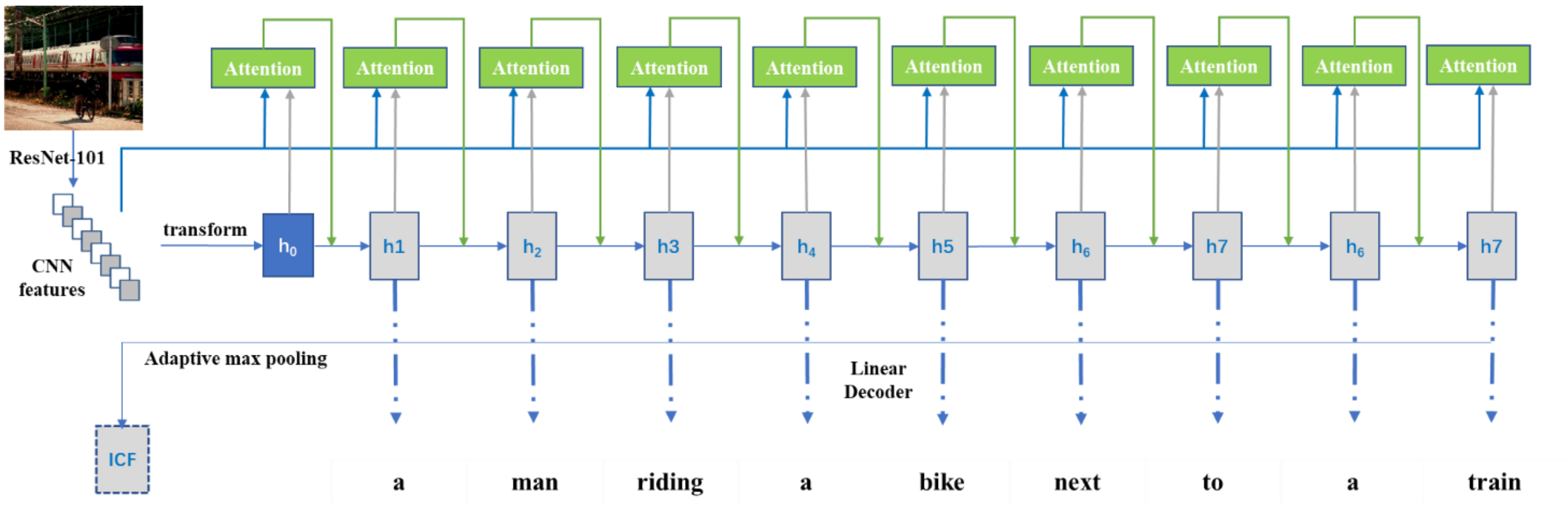

Figure 2. ICF based on ICM and adaptive max pooling. ICM inputs the CNN features using ResNet-101, and uses RNN module to abstract and obtain more semantic features. CNN features, attention module and previous state $h_{n\text{-}1}$ are considered to produce the next state $h_n$. These gray rectangle with solid line represents the middle state $h_n$ in RNN module. Each middle state can be transformed into one word through the linear decoder. From the view of these words, these states $h_n$ contain features about objects ("man", "train", "bike"), quantity ("a" man, "a" train), and relationships ("riding", "next to") in the image, which represents the global semantic understanding of the image. Based on these states $h_n$, the ICF (gray rectangle with dotted line) of the image can be obtained through adaptive max polling, which solve the problem of quantity inconsistency for different lengths of words captions.

As shown in the Figure 2, the CNN features extracted through ResNet-101 were transformed as the initial input state $h_0$ of RNN module. At each step, the decoder predicts next state $h_n$ that is a fixed dimensionality (512-D) of features, by using the previously generated 512-D features ($h_{n\text{-}1}$) and CNN features of the image. Then a linear layer is used as decoder to transform the predicted 512-D features ($h_n$) into a probability vector, of which each value accords with each word in the vocabulary, hence the corresponding word with the maximal value is obtained. For the caption of one image, each word was predicted based on different part of the input image. For example, the prediction of the word "man" is only related with those pixels that represent the man. Hence, the attention network was added to predict where the RNN module needs to pay more attention for entire image space when predicting next state $h_{n+1}$. The attention module input the CNN features of the image and the previous state $h_{n\text{-}1}$, then output the weighted CNN features to predict the next state $h_n$. More details about ICM can be referred to [35], the pre-trained ICM can be downloaded from https://github.com/sgrvinod/a-PyTorch-Tutorial-to-Image-Captioning.

Next, we constructed ICF of images based on these middle state $h_n$ (512-D features). Because each image may generate different length of words based on ICM, which render $n$ different for different images. Hence, different number of 512-D features will be generated. However, the following linear regression model requires the fixed dimension of input. In order to solve the problem, we employed the adaptive pooling across the different number of 512-D features. We employed the maximize operation for each value in 512-D vector, and we can obtain one 512-D features regardless of how many 512-D features. The final 512-D feature vector is the ICF of the image for subsequent encoding.

**Linear regression mapping from ICF to voxels.**

For each voxel in visual cortices, one linear regression model is used to map ICF to voxel. In this study, we assume that each voxel can be characterized by small part of features and regularize the linear mapping sparse to prevent overfitting when training each linear regression model. We employ the regularized orthogonal matching pursuit (ROMP) [36] to solve the linear regression problem. ROMP adds an orthogonal item and group regularization based on the matching pursuit (MP) algorithm [37]. In this way, we trained these linear models in voxel-wise manner. The voxels can be predicted based on ICFVEM including ICF extracting module and linear regression models, given one new image stimulus of testing set. In the same way, we constructed voxel-wise encoding models using each of 35 selected layers of features from 101 layers of encoder as the control models. Thus, the corresponding encoding performance and comparison of traditional CNN features for classification and ICF for image caption can be calculated to validate our proposed model.

## Results

**Hierarchical information processing for deep network and human visual cortices.**

Firstly, we presented the encoding performance of different layers of ResNet-101 to validate the effectivity of the deep network architecture. For each voxel, we calculated Pearson correlation (PC) between true and predicted responses for testing set, and computed the average PC for all voxels in specific visual cortices to evaluate encoding performance. As shown in the Figure 3, we combined four high-level visual cortices into one visual cortices, and presented the encoding performance of four visual cortices including left and right low-level and high-level visual cortices. We can see that roughly rising trend for left and right high-level visual cortices and firstly rising and then descending trend for low-level visual cortices. The appearance can be also seen in the previous studies about visual encoding based on deep network models. Deep network models perform hierarchical representation of input images and obtain hierarchical features ranging from low-level to high-level visual cortices, which is similar with the hierarchical processing of visual information in human visual cortices. In human visual system, visual information flows from low-level visual cortices to high-level visual cortices and human understands the external scene through the hierarchical processing. The characteristic makes it valuable to analyze human visual processing with the deep network architecture that is currently regarded as the best of brain-like structures. In this study, we are more interested in how to further improve the encoding performance of high-level visual cortices, after using the last few layers of features. From the perspective of driven task, we analyzed the drawbacks of previous deep network models in the introduction section, and we introduced the IC task to design ICFVEM for high-level visual cortices, which expands the horizon of current visual encoding.

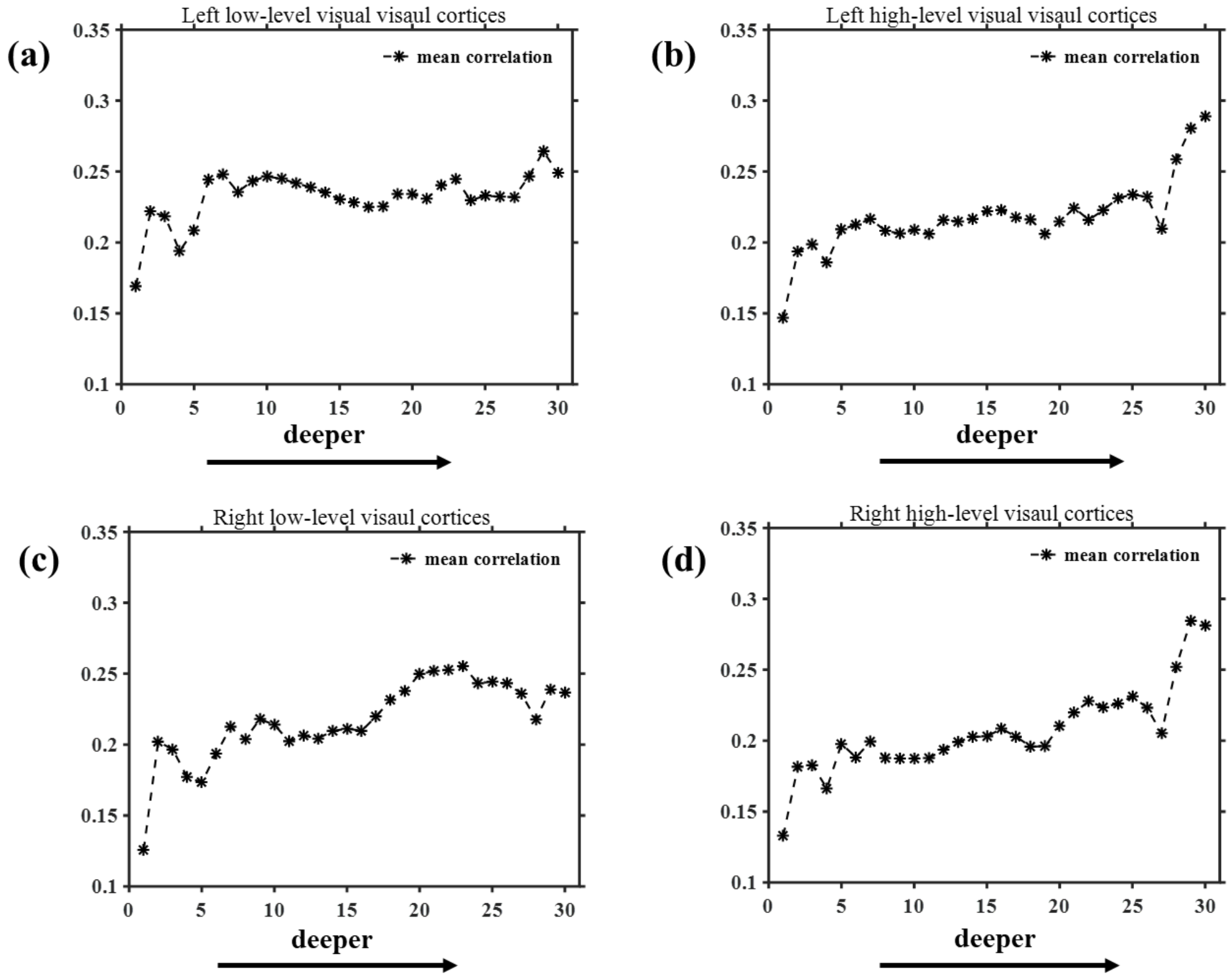

Figure 3. Hierarchical encoding performance based on hierarchical features of the deep network model. In total, 30 features ranging from front layers to last layers are used to encode the voxels of the four visual cortices including left and right low-level and high-level visual cortices. Each layer of features can be used to encode each visual cortices, and mean correlation is used to evaluate the encoding performance of current layers of features for specific visual cortices. As the features get deeper, the encoding performance behaves better for high-level visual cortices, and firstly rising and then descending for low-level visual cortices. These trends can reflect that the hierarchical features in deep network model, and the hierarchical visual information processing in visual cortices.

### Comparison between ICF and CNN features for visual encoding.

In the experiment, the model based on CNN features for visual encoding is called CNNM in short. For 30 selected CNN layers, we selected the best layer as the control model according to encoding performance for each of 10 sub-regions. To compare the encoding performance of two models (CNNM and ICFVEM), we firstly made a scatter plot in the Figure 4, where each dot corresponds to a single voxel, and the ordinate and abscissa of each dot represents PC values of CNNM and ICFVEM, respectively. Secondly, in the Figure 5, we plotted the distribution of encoding performance difference of the voxels on whom both models yielded significant prediction. Note that the threshold of significant PC value is 0.27 ($p<0.001$). Results for three subjects can be seen in Figure 4 and 5, and we presented the encoding performance for left (L) and right (R) low (L) and high (H) -level visual cortices, which join into four regions (LL, LH, RL, and RH).

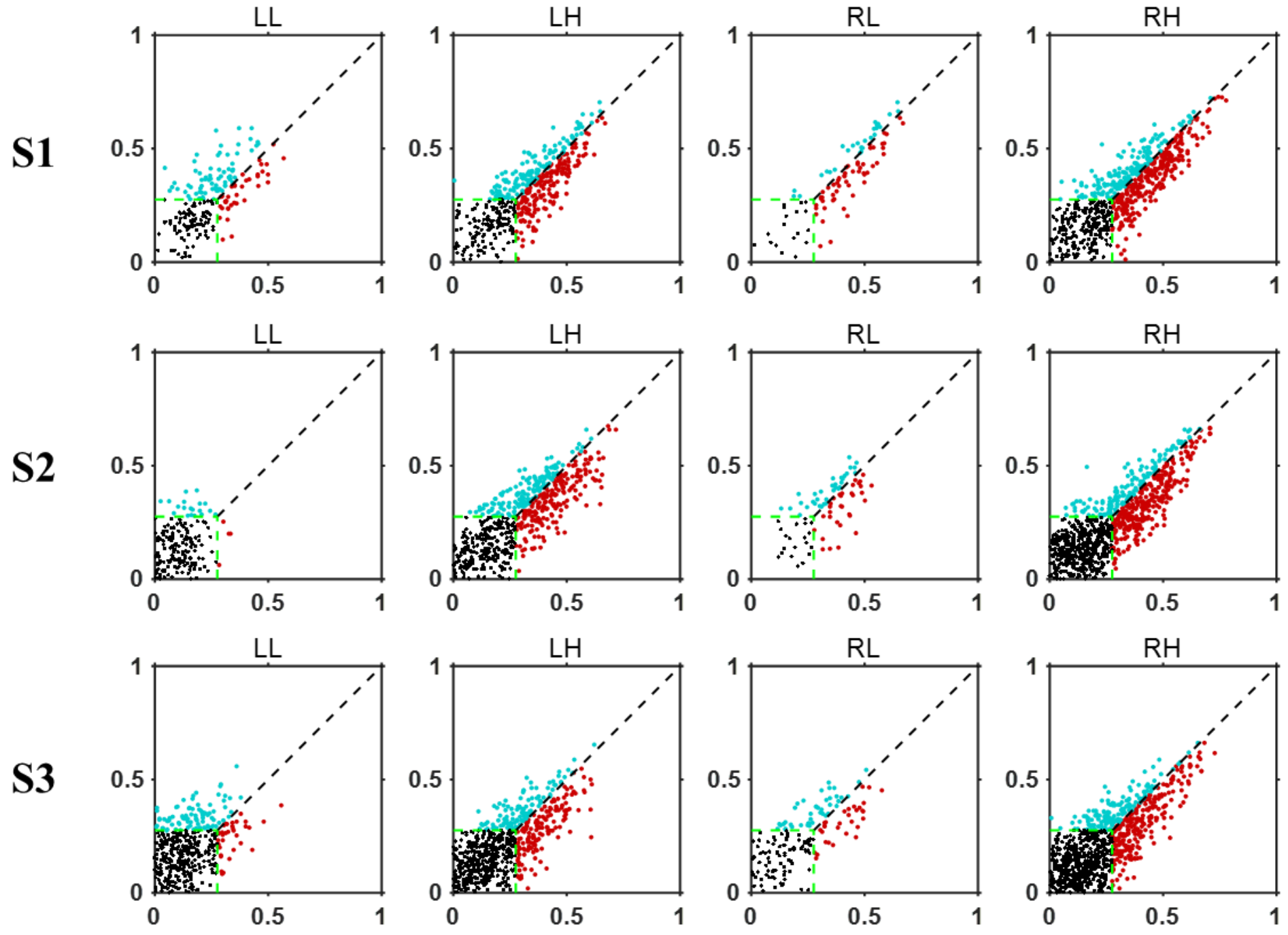

Figure 4. Encoding performance of ICFVEM compared to CNNM for three subjects from the perspective of quantity. Each subfigure presents a comparison of the two models in a specific visual ROI for specific subject. In each subfigure, the ordinate and abscissa of each dot represent the PC values of CNNM and ICFVEM respectively. The title “LL”, “LH”, “RL”, and “RH” represents the left low-level, left high-level, right low-level, and right high-level visual cortices. The green dashed lines represent the threshold (0.27, $p < 0.001$) for significant PC value, hence, the black dots correspond to those voxels cannot be effectively encoded (under 0.27) by either of CNNM and ICFVEM. The red dots correspond to those voxels that can be better encoded by ICFVEM than CNNM and vice versa for the cyan dots.

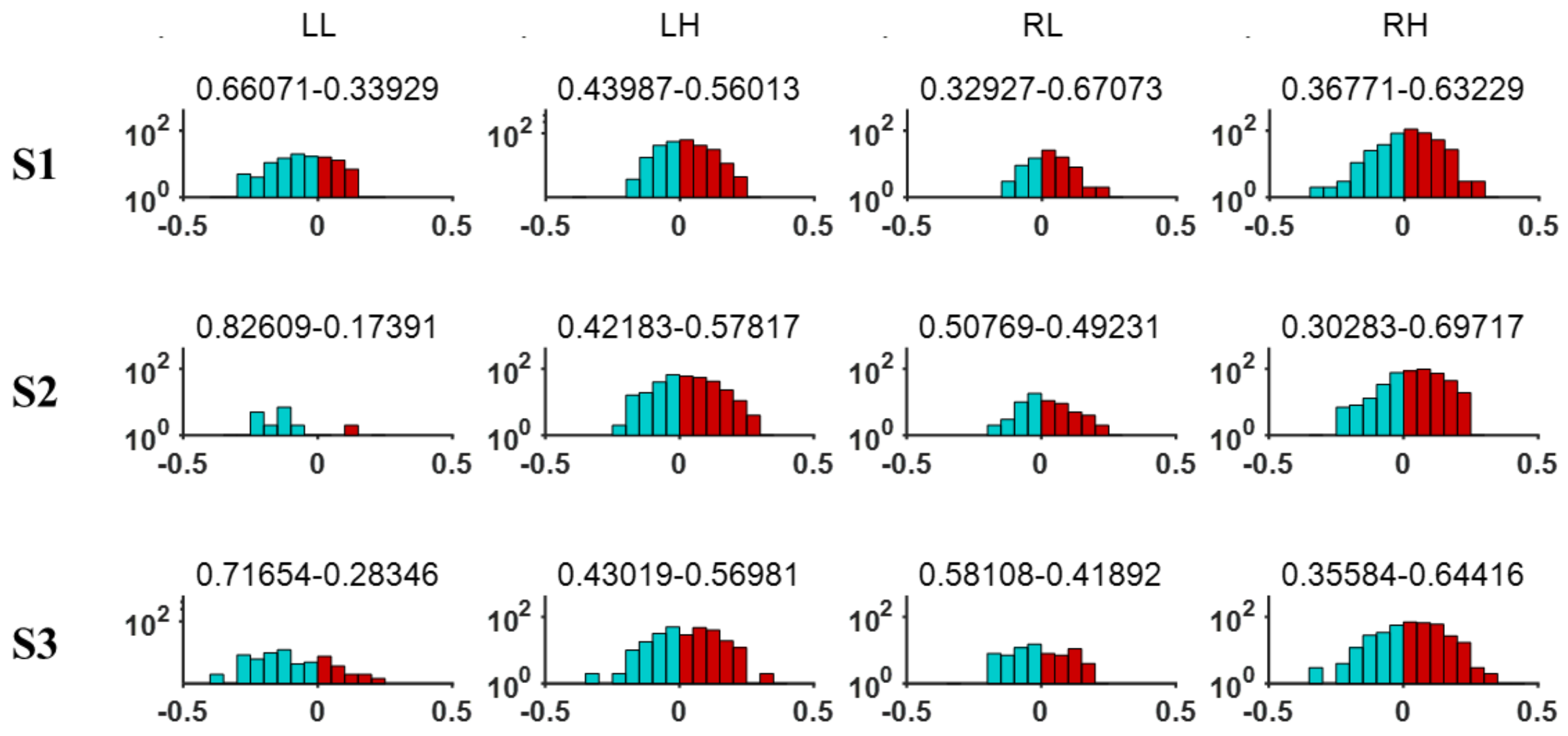

Figure 5. Encoding performance of ICFVEM compared to CNNM for three subjects from the perspective of distribution. Those pillars above 0 indicate higher prediction accuracy for ICFVEM, as marked by red color, and vice versa for the cyan color for GWPM. The number for each side represents the fraction of voxels whose encoding performance are higher than that of the other model.

In comparison, we can see that the encoding performance for ICFVEM aimed at high-level semantic task is better in left and right high-level visual cortices, although ICFVEM behaves slightly worse in low-level visual cortices. For example, in Figure 4 for high-level visual cortices, the number of those red dots that represent ICFVEM is more than those cyan dots that represent CNNM, which indicates that ICFVEM have higher encoding accuracy than CNNM. In Figure 5 for high-level visual cortices, the occupancy rate of the red pillars that represents ICFVEM is 15%-30% higher than cyan pillars that represents CNNM. In addition, the performance of the three subjects in Figure 4 and 5 behaves the similar results uniformly, which validates ICFVEM across the subjects and indicates that the proposed ICFVEM is robust. In this way, ICF behaves more semantic based on CNN features, which demonstrated that the deep network can be driven by the more semantic task and perform more semantic feature representation.

ICFVEM gives rise to the consideration of task used to drive deep network models in visual encoding domain. As shown in the Figure 1, the participant subject in the fMRI experiment did not only know the main object like the image classification, but really understand what he or she viewed. In this way, the deep network models driven by image classification are not enough for visual encoding especially high-level visual cortices, the deep network architecture need to be driven by more semantic task like the participant subjects. Apparently, IC task approach to it closer than image classification task, and the results also validate ICFVEM and the original assumption: target task also plays an important, in addition to deep network architecture.

Further, we presented the detailed encoding performance of sub-regions for four high-level visual regions: LOC, OPA, PPA, and RSC in the Figure 6. For each sub-region, we used the PC values of all voxels based on ICF and best layer of features, respectively, to compute the average absolute distance to evaluate the difference of representations. We can see that ICF and CNN features demonstrated more difference for L-LOC, R-LOC, ROPA, and R-PPA across the three subjects. Except main object for image classification, IC task pay additional attention to other objects and relationships between objects. Hence, it can be indicated that L-LOC, R-LOC, and R-PPA also focus on the representations about other local objects or relationships between objects, except the main objects. In contrast, L-OPA and L-RSC behaves less difference compared to the other regions, which implies that these regions tends to be careful for main object of image stimuli. Certainly, some regions demonstrated different characteristics for difference subjects and may have personalized representations, such as L-PPA and R-RSC.

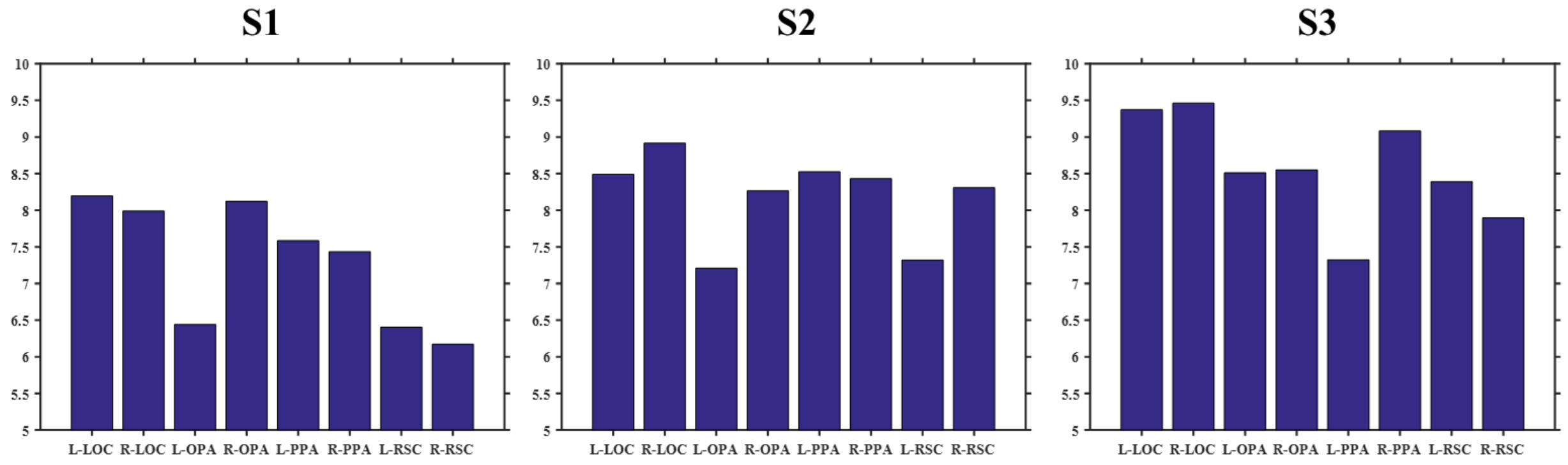

Figure 6. Encoding difference of ICFVEM and CNNM for different high-level visual regions. Three subjects are all presented and each subfigure accords to one subject. In each subfigure, the ordinate and abscissa represent the average distance values and eight visual sub-regions. The average distance values (ordinate) are used to evaluate the difference of representation of image stimuli for each sub-region from the perspective of features, because ICF and CNN features have different representations.

**Interpretation of voxels based on different type of words**

Moreover, we explore those voxels well-matched with ICF and analyzed the semantic function of these voxels through some words related with ICF. As ICF is obtained through the adaptive max pooling of several 512-D features, each of which corresponds to one word, we constructed the connection between voxels and words. Based on the linear regression mapping of ICFVEM, we employed each 512-D features to predict each of those significant voxels whose PC values are higher than significance threshold (0.27). In this way, we can determine those words related with each voxel according to the predicting accuracy. For each voxel, we selected two words for each image stimulus in testing set. In Figure 7, we plotted the distribution of words for each voxel based on the wordcloud package that can be downloaded from https://github.com/amueller/word_cloud.

Those words that appear more times are bigger, and vice versa. We can conclude some rules about some voxels, and the interpretation is applicable to different visual regions and subjects. For example, 138$^{th}$ voxel in L-LOC of the first subject demonstrated more relationship with the description of spatial position, such as “next”, “close”. 20$^{th}$ voxel in R-RSC of the second subject is related with “man” and “people”. 17$^{th}$ voxel in R-RSC of the third subject is related with the color (“white” and “red”). 89$^{th}$ voxel in L-PPA of the third subject and 143th voxel in L-LOC of the first subject tend to be the representation of quantity (“group” and “large”). The interpretation of these voxels reflects semantic characteristics, which cannot be obtained through previous deep network models pre-trained on image classification, and validates the advantage of ICFVEM. In this way, the relationship between voxels and semantic words can be revealed.

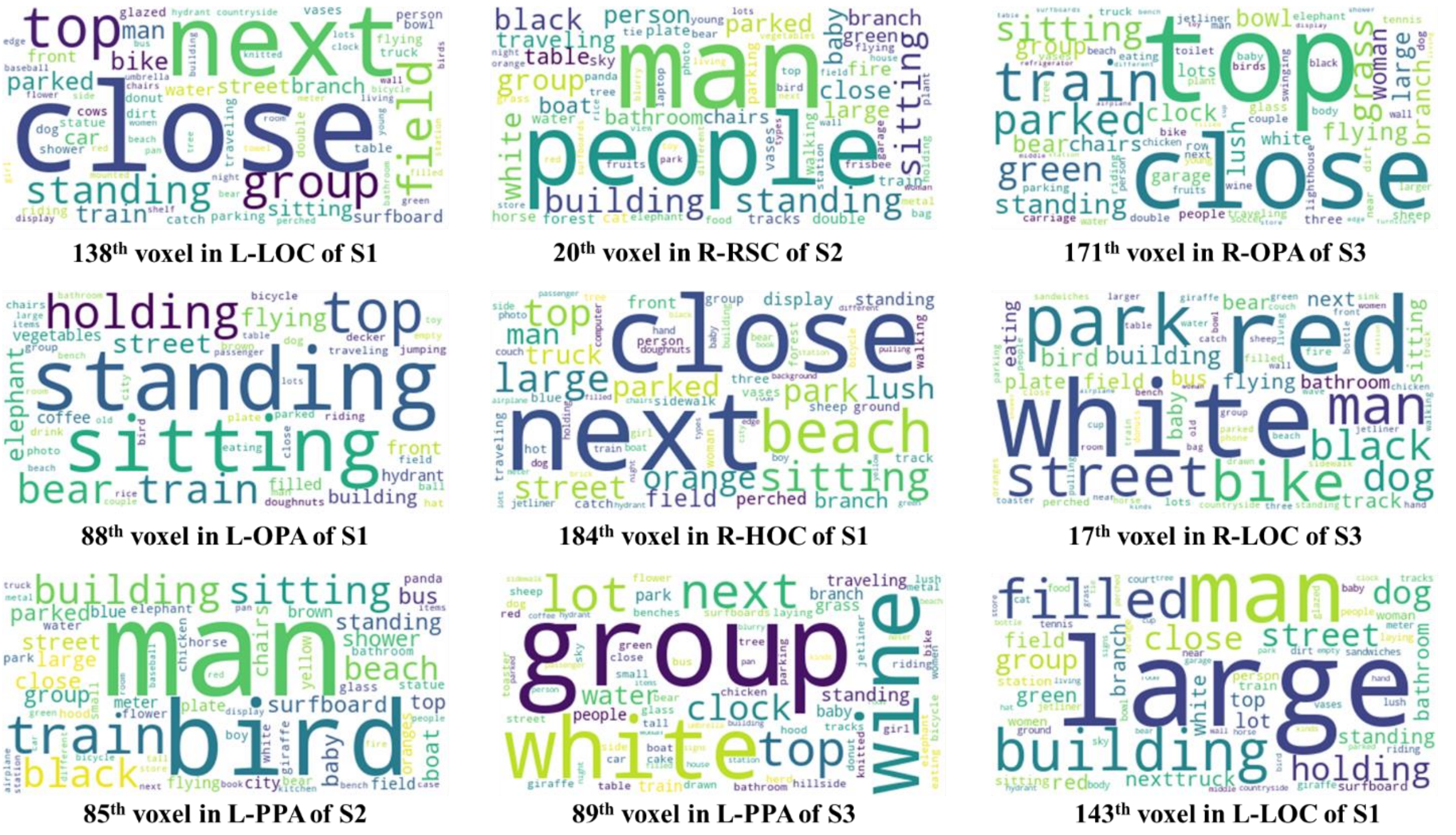

Figure 7. Interpretation of some typical voxels based on related words. One subfigure accord to one voxel in specific high-level visual region for one subject. Different words are presented using different color. The more times the word appears, the bigger space the word occupy. The statistical property of words for different voxels can be revealed through the visualization.

Further, we use the interpretation of some voxels to explore the relationship some voxels of different visual regions and subjects. As shown in the Figure 8, we can see that some voxels can represents the same and different representations in the same visual region such as R-LOC, which implies that the representations of voxels in one visual region are overlapping and complementary . In the Figure 9, there exist the some voxels that represents similarly in different visual regions, which indicates that the representation of different visual regions is not completely different but overlapping. In the Figure 10, some voxels from different subjects represents similarly, which can be help design encoding or decoding models across subject. The above detailed interpretation enhanced the understanding the characteristics of voxels, which benefit from the ICFVEM and the visualization based on related words. In addition, the relationship between voxels and words cannot be comprehensively covered, because the number of image stimuli is limited and in not all words in dictionary are used. In future, big dataset of fMRI can further promote ICFVEM and comprehensive interpretation of voxels, through which, detailed semantic characteristics can be revealed.

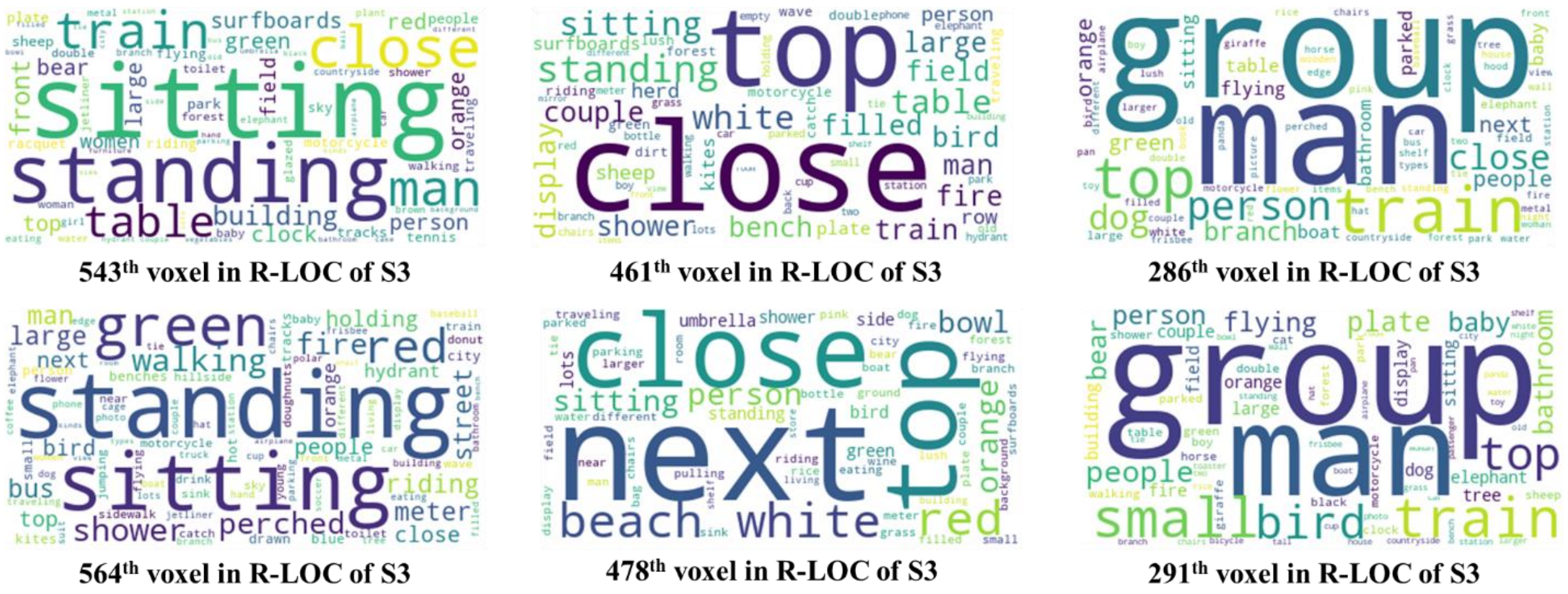

Figure 8. Interpretation of some voxels with the same and different representations in one visual region. Each column represents two subfigures that have similar representations. Each row represents three subfigures that have different representations.

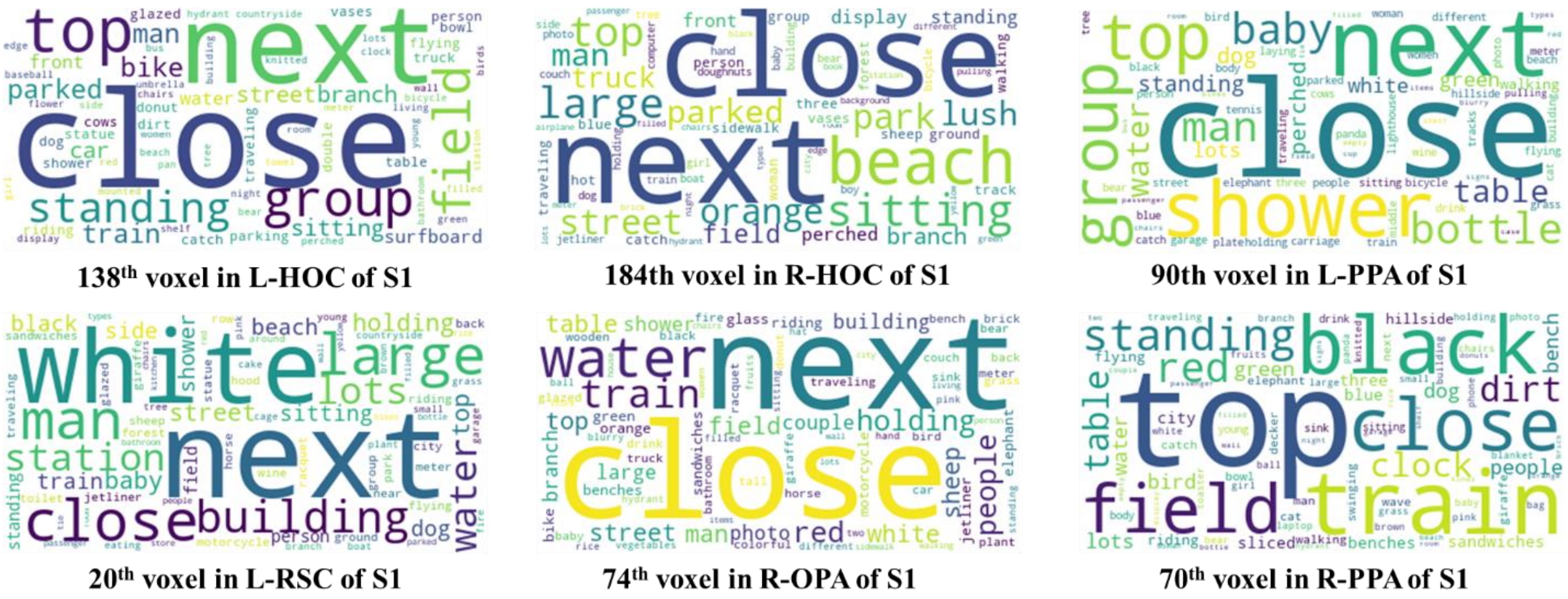

Figure 9. Interpretation of some voxels with the similar representations for different visual regions. The six subfigure demonstrated similar representations about the description of spatial relation.

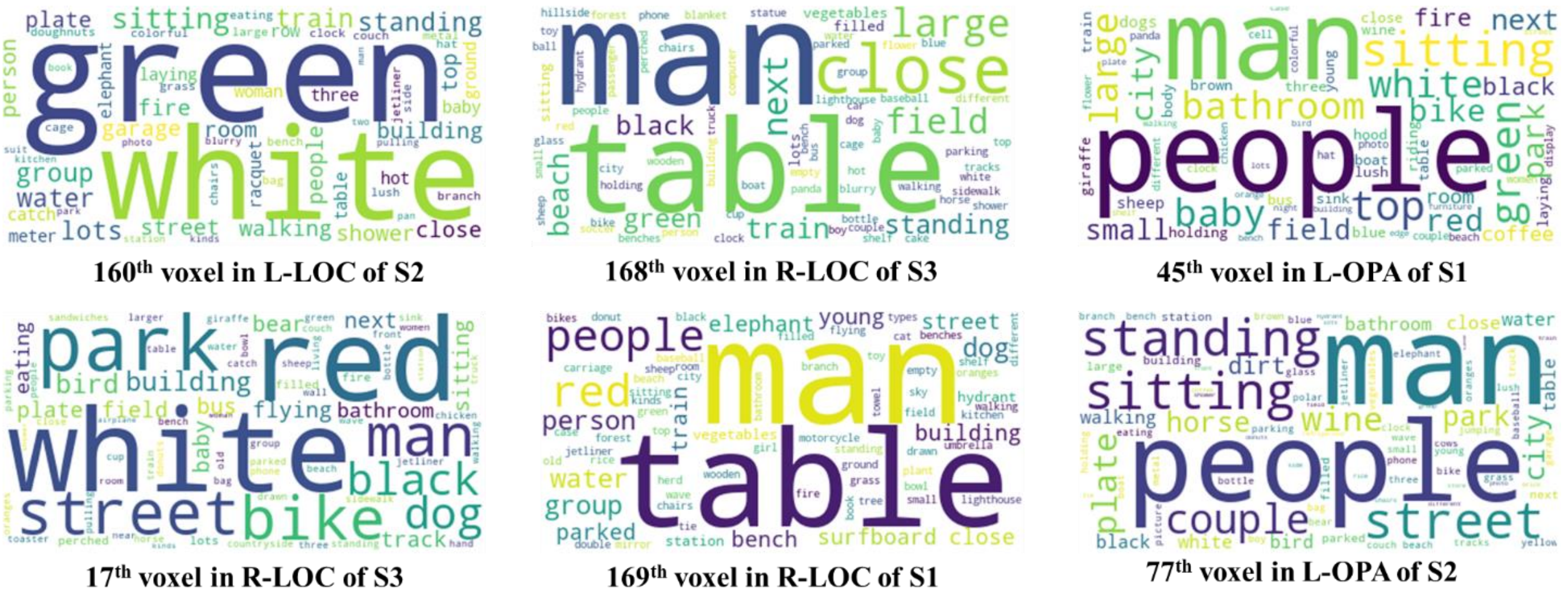

Figure 10. Interpretation of some voxels with the same representations for different subjects. Subfigures in each column have similar representations across different subjects.

# Discussion

## Correlative representation in high-level visual cortices

ICM can extract those features about the relationship between objects in images. Those deep network models pre-trained on image classification task are hard to extract these features to characterize the relationship, because image classification task is more care of the main object for invariance. The encoding performance of high-level visual cortices was improved through introducing ICF, and the interpretation of voxels using word visualization also demonstrated that some voxels are well related with the description of “quantity”, “connection”, or “location”, which can be seen in the Figure 7. Hence, the improvement and interpretation are partly due to the correlative representation including all objects and their relationship for input images used to form the global understanding of images. In this way, it implies that high-level visual cortices perform more semantic correlative representation of images, when viewing external images for subjects.

## Correlative consideration between computer vision and human vision

As shown in the Figure 1, deep network features of image stimuli are used to predict fMRI patterns of participant subjects. ICFVEM gave rise to the consideration of task used to drive deep network models and the improved results validated the original viewpoint: target task in computer vision plays an important in designing the visual encoding model. For more comprehensive consideration, it needs to take into account of other aspects of computer vision and human vision domain, before designing visual encoding models. In computer vision domain, task, architectures of deep network models, training techniques, etc., have an influence on the feature representation of trained deep network models. In human vision, image stimuli, task when viewing images for subject can influence neuron activity and the corresponding fMRI pattern. However, current studies for visual encoding confine to the deep network pre-trained on image classification task. Considering that these elements are all important for visual encoding, detailed analyzation and correlative consideration between computer vision and human vision may be the future direction for visual encoding.

## Conclusions

By analyzing the influence of the driven task of deep network model on visual encoding, this study gave the novel viewpoint: the previous deep network models pre-trained on classification task are not enough for the encoding of high-level visual cortices. Moreover, we introduced ICM aimed at another more semantic task: image caption, and proposed ICFVEM to encode voxels of high-level visual cortices. Experiment on three subjects also demonstrated that more semantic features can contribute more to high-level visual cortices. In addition, the ICF implies that the high-level visual cortices perform the correlative representation. The visualization of voxels using words help understand the detailed semantic characteristics of voxels in high-level visual cortices. In the future, other elements related to deep learning including task designing, training technique, and so on, may be paid more attention to design the computation models for visual encoding, for the sake of deep understanding of human visual information processing.

## Data Availability

The detailed information and downloaded address about the used fMRI data [34] can be found in *https://figshare.com/articles/BOLD51200/6459449*.

## Conflicts of Interest

There is no conflict of interest for the authors regarding the publication of this paper.

## Funding Statement

This work was supported by the National Key Research and Development Plan of China (No. 2017YFB10025122), the National Natural Science Foundation of China (No. 61701089), and the Natural Science Foundation of Henan Province of China (No. 162300410333).